\pdfoutput=1

\documentclass[11pt]{article}

\usepackage[preprint]{acl}
\usepackage{booktabs}
\usepackage{times}
\usepackage{latexsym}

\usepackage[T1]{fontenc}

\usepackage[utf8]{inputenc}

\usepackage{color, colortbl}
\definecolor{deepred}{rgb}{0.631,0.102,0.102}
\definecolor{gyellow}{HTML}{F4B400}
\definecolor{mildyellow}{HTML}{FFF2CC}

\definecolor{deepblue}{RGB}{0,102,204}
\definecolor{lightblue}{RGB}{173, 216, 230}
\definecolor{deeporange}{RGB}{255, 140, 0} 

\newtcolorbox{Case Study with Fixed-ratio Inappropriate Context}{
  colback=white,
  colframe=deepblue,
  fonttitle=\bfseries,
  title= Prompt for Claim Generation,
  rounded corners,
}

\newtcolorbox{Case Study with Fixed-ratio Inappropriate Context1}{
  colback=white,
  colframe=orange,
  fonttitle=\bfseries,
  title= Prompt for Generating Atomic Reasoning Chain,
  rounded corners,
}

\newtcolorbox{Case Study 2 on Randomized Proportions and Order of Inappropriate Context}{
  breakable,
  colback=white,
  colframe=lightblue,
  fonttitle=\bfseries,
  title= Disproportionate Inappropriate Context,
  rounded corners,
}

\usepackage{microtype}

\usepackage{listings}

\usepackage{inconsolata}

\usepackage{graphicx}
\usepackage{multirow}
\usepackage{amsmath}
\usepackage{amssymb} 
\usepackage{tabularx}
\usepackage{arydshln}

\usepackage{pifont}

\definecolor{newbleu}{HTML}{d0e0ee}

\definecolor{customgreen}{HTML}{16C47F}  
\definecolor{customred}{HTML}{C62300}   

\newcommand{\cmark}{\textcolor{customgreen}{\ding{51}}}  
\newcommand{\xmark}{\textcolor{customred}{\ding{55}}}   
\definecolor{darkblue}{rgb}{0, 0, 0.5}
\hypersetup{colorlinks=true, citecolor=darkblue, linkcolor=darkblue, urlcolor=darkblue}

\NewDocumentCommand{\heng}
{ mO{} }{\textcolor{red}{\textsuperscript{\textit{Heng}}\textsf{\textbf{\small[#1]}}}}

\NewDocumentCommand{\chenkai}{ mO{} }{\textcolor{teal}{\textsuperscript{\textit{Chenkai}}\textsf{\textbf{\small[#1]}}}}

\NewDocumentCommand{\yuji}{ mO{} }{\textcolor{blue}{\textsuperscript{\textit{Yuji}}\textsf{\textbf{\small[#1]}}}}

\NewDocumentCommand{\qingyun}{ mO{} }{\textcolor{orange}{\textsuperscript{\textit{Qingyun}}\textsf{\textbf{\small[#1]}}}}

\NewDocumentCommand{\jiateng}{ mO{} }{\textcolor{green}{\textsuperscript{\textit{jiateng}}\textsf{\textbf{\small[#1]}}}}

\NewDocumentCommand{\cheng}{ mO{} }{\textcolor{purple}{\textsuperscript{\textit{Cheng}}\textsf{\textbf{\small[#1]}}}}

\NewDocumentCommand{\zdh}
{ mO{} }{\textcolor{red}{\textsuperscript{\textit{Denghui}}\textsf{\textbf{\small[#1]}}}}

\NewDocumentCommand{\preslav}
{ mO{} }{\textcolor{orange}{\textsuperscript{\textit{Preslav}}\textsf{\textbf{\small[#1]}}}}

\title{\includegraphics[width=0.039\linewidth]{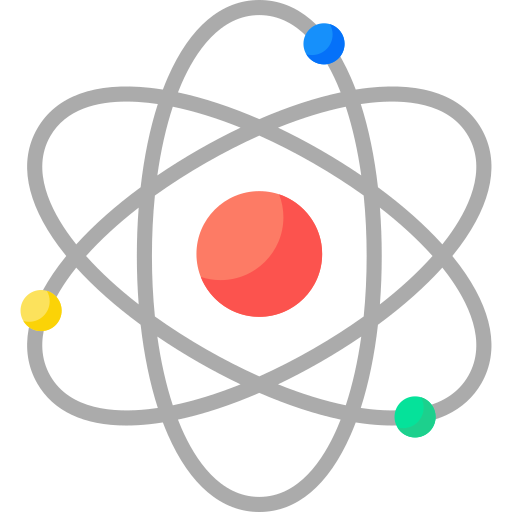} 
Atomic Reasoning for Scientific Table Claim Verification}

\author{
Yuji Zhang$^{1}$,  Qingyun Wang$^{1}$, Cheng Qian$^{1}$, Jiateng Liu$^{1}$, Chenkai Sun$^{1}$, Denghui Zhang$^{1, 2}$\\
\textbf{Tarek Abdelzaher$^{1}$, Chengxiang Zhai$^{1}$, Preslav Nakov$^{3}$, Heng Ji$^{1}$}\\
$^{1}$University of Illinois Urbana-Champaign\\$^{2}$Stevens Institute of Technology, $^{3}$MBZUAI\\
\texttt{\{yujiz, hengji\}@illinois.edu}\\
}

\begin{document}
\maketitle
\begin{abstract}

Scientific texts often convey authority due to their technical language and complex data. However, this complexity can sometimes lead to the spread of misinformation. Non-experts are particularly susceptible to misleading claims based on scientific tables due to their high information density and perceived credibility.
Existing table claim verification models, including state-of-the-art large language models (LLMs), often struggle with precise fine-grained reasoning, resulting in errors and a lack of precision in verifying scientific claims.
Inspired by Cognitive Load Theory, we propose that enhancing a model's ability to interpret table-based claims involves reducing cognitive load by developing modular, reusable reasoning components (i.e., \textbf{atomic skills}). We introduce a skill-chaining schema that dynamically composes these skills to facilitate more accurate and generalizable reasoning with a reduced cognitive load. 
To evaluate this, we create SciAtomicBench\footnote{Our datasets and models will be released upon publication at \url{https://github.com/CelestineZYJ/SciAtomicBench}.}, a cross-domain benchmark with fine-grained reasoning annotations.
With only 350 fine-tuning examples, our model trained by atomic reasoning outperforms GPT-4o's chain-of-thought method, achieving state-of-the-art results with far less training data.

\end{abstract}

\section{Introduction}






\begin{figure*}[htb!]
\vspace{-2em}
\centering
\includegraphics[width=1.0\linewidth]{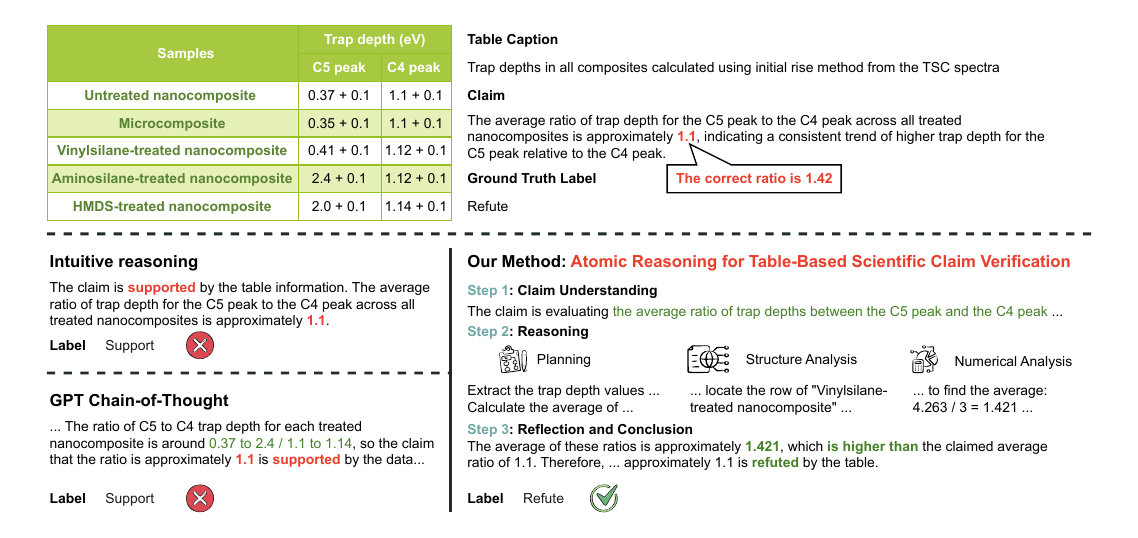}
\vspace{-2em}
\caption{\label{fig_table:overview} A material science table claim verification example from our SciAtomic benchmark illustrates the difference between intuitive reasoning, CoT reasoning, and our proposed atomic reasoning. }
\vspace{-1em}
\end{figure*}

In scientific domains, specialized terminology, complex presentation, and the aura of expertise lend information authority but also make it vulnerable to intentional distortion and rapid propagation of misinformation~\cite{cabanac2021tortured,else2021tortured,lim2021you}. Non-expert readers, lacking deep domain knowledge, are susceptible to accepting misleading or erroneous claims as fact due to the inherent credibility carried by scientific content~\cite{osborne2023science}. Such misplaced trust can have serious consequences; for instance, fraudulent heart stem cell research misled scientists, leading patients to undergo ineffective treatments and diverting resources from legitimate medical advancements~\cite{kowalczyk2017partners}. Specifically, the CONCERT-HF trial, based on the fraudulent heart stem cell research, resulted in patients' death~\cite{bolli2018rationale}. Consequently, detecting and flagging scientific misinformation is both urgent and essential.

In the scientific domain, tables serve as a crucial medium for recording and representing data by condensing complex data into an accessible format~\citep{inskip2017getting}, and many claims hinge on their precise interpretation. Yet, non-experts often struggle to parse these dense formats, making tables a prime vector for subtle misrepresentation. 
Therefore, an accurate understanding of tables is crucial for validating the factuality of corresponding claims in papers and maintaining a trustworthy scientific environment. To address this, researchers have used large language models (LLMs) to tackle scientific table fact-checking challenges by verifying claims against tables~\citep{gupta-etal-2020-infotabs, Chen2020TabFact,wang-etal-2021-semeval,akhtar-etal-2022-pubhealthtab,lu-etal-2023-scitab}.
However, models tailored to tables still fall short when applied to the nuanced demands of scientific domains~\citep{lu-etal-2023-scitab}. Additionally, even state-of-the-art closed-source LLMs like GPT-4~\cite{openai2023gpt4}, achieve worse than human performance by 20\%~\cite{lu-etal-2023-scitab}.

Our analysis reveals a key limitation of existing models: they often lack the awareness to explicitly decompose and invoke fine-grained atomic skills during reasoning, which limits their generalization to complex table comprehension tasks.
As shown in Figure~\ref{fig_table:overview}, our atomic reasoning method breaks verification into three essential skills in fine-grained reasoning steps: (1) conceptual matching, (2) value extraction, and (3) numerical calculation. When these skills are explicitly applied, the model arrives at the correct conclusion.
In contrast, ChatGPT, whether via intuitive or Chain-of-Thoughts (CoT)~\cite{wei2022chain} reasoning, fails to isolate these operations, instead relying on reasoning that mixes or skips steps, leading to errors.

Inspired by Cognitive Load Theory~\cite{plass2010cognitive}, which posits that human working memory is limited and that learning is optimized when instructional design minimizes extraneous burden, we observe that large language models (LLMs) face similar challenges when interpreting information-dense scientific tables. Since models cannot flexibly identify and reuse the common reasoning operations underlying a large and diverse set of table claims, they incur a heavy cognitive load and processing burden. To address this, we introduce a suite of highly modular, reusable, and generalizable competencies, termed \textbf{atomic skills}, each encapsulating a distinct reasoning operation (e.g., concept disambiguation, numerical calculation, trend check). Decomposing the heavy verification workflow into these atomic skills reduces the model's inference burden and promotes stronger generalization across novel table and claim types and diverse domains. 

\begin{figure*}[htb!]
\vspace{-2em}
\centering
\includegraphics[width=\linewidth]{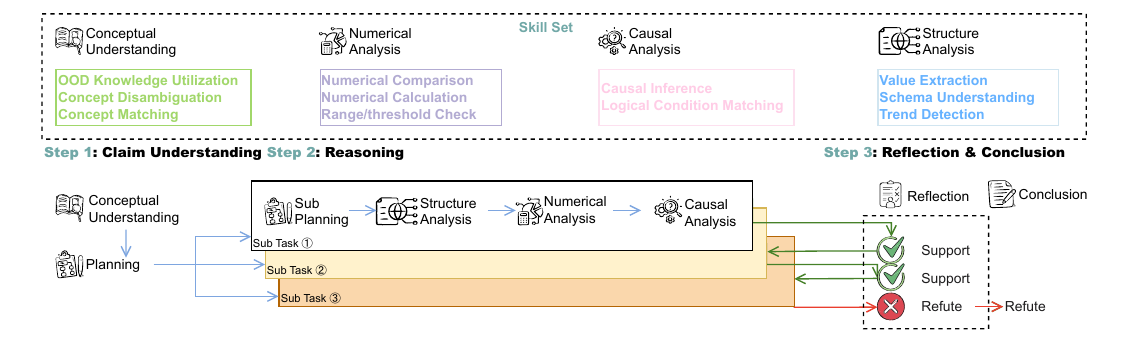}
\vspace{-2em}
\caption{\label{fig_table:pipeline} Illustration of our skill-chaining schema with fine-grained reasoning steps equipped with atomic skills.}
\end{figure*}

To equip models with atomic skills, we curate a compact, diverse dataset for capability training and introduce a skill-chaining schema that generates skills in context, as shown in Figure~\ref{fig_table:pipeline}. Traditional CoT prompting decomposes a task into a monolithic reasoning path~\cite{wei2022chain}, neither enforcing granular skill control nor preventing information loss in lengthy sequences, an issue shown to degrade performance when critical details vanish mid-chain~\cite{liu-etal-2024-lost}. Our schema instead defines precise context and goals for each step and links them so that each consumes only its predecessor's output. Under these adequate conditions, models dynamically select and pack the appropriate atomic skills, learning when and how to invoke each modular competency rather than indiscriminately scaling up chain length. 
This disciplined, modular approach reduces cognitive burden, leverages inference-time scaling more effectively, and fosters generalization to diverse table claims.

In our pursuit of more enhanced table claim interpretation, we address a second critical challenge: existing scientific table claim verification benchmarks suffer from limited domain diversity and insufficient complexity. Expert‐annotated datasets are inherently sparse, and many benchmarks include claims that fail to mirror the complex nature of real‐world scientific inquiry. To overcome these limitations, we introduce a comprehensively curated, cross‐domain benchmark spanning material science, medicine, finance, and machine learning. 
Our dataset includes detailed annotations of atomic reasoning chains and provides a balanced distribution of claim difficulty and reasoning complexity. 

Our experimental results demonstrate the strong efficiency and effectiveness of our AtomicTableLLM. With only 350 fine-tuning examples, we boost the performance of the Deepseek-Qwen-7b-based model from 63.12\% to 85.70\% on the Finance domain.
On the public SciTab benchmark, our model outperforms GPT-4o with CoT reasoning, and surpasses state-of-the-art baselines, which are typically trained on much larger datasets (e.g., million-size).
Our contributions are threefold:

\begin{itemize}
    \item We propose data-efficient atomic reasoning, enabling language models to learn highly modular and composable reasoning skills that enhance their generalization across claim types and scientific domains.
    \item We develop table-specialized LLMs, AtomicTableLLM, which surpasses state-of-the-art models in scientific table-claim verification, demonstrating superior reasoning and generalization capabilities across scientific domains.
    \item We construct a new scientific table-claim dataset, \textbf{SciAtomicBench}, annotated with fine-grained atomic skills and long reasoning chains, covering multiple scientific domains, including material science, medical science, finance, and computer science.

\end{itemize}

\section{Related Work}

\subsection{Table-Specialized Large Language Models}

Earlier work on table language models focused on resources in the general domain~\citep{Lehmberg2016WebTable,wang-etal-2018-describing,hu2019viznet,yin-etal-2020-tabert,deng2020turl,herzig-etal-2020-tapas,iida-etal-2021-tabbie,wang-etal-2021-stage,xie-etal-2022-unifiedskg,liu2022tapex}.
Recent advancements in large language models~\citep{dinh2022lift,pmlr-v206-hegselmann23a,jiang-etal-2023-structgpt,chen-2023-large,zhao-etal-2023-large,li2023sheetcopilot,li2024tablegpt,zhang-etal-2024-e5} have shown impressive zero-shot and few-shot performance in table understanding tasks. To further improve reasoning ability and performance, other large language model (LLM) techniques have been applied to table-specialized LLMs, including data augmentation~\citep{li2024tablegpt,zhang2024tablellm}, instruction tuning~\citep{pmlr-v206-hegselmann23a,zhang-etal-2024-tablellama}, prompt engineering~\citep{jiang-etal-2023-structgpt,deng-etal-2025-enhancing}, in-context learning~\citep{zhao-etal-2023-large}, code generation~\cite{lu-etal-2025-tart,zhang-etal-2025-alter}, chain-of-thought reasoning~\citep{chen-2023-large,zhang-etal-2024-e5}, and multi-agent collaboration~\citep{li2023sheetcopilot}. Despite the promising progress in the general domain, no prior work focuses on scientific papers, for which the availability of high-quality annotation data is limited. Moreover, we are the first to decompose the table reasoning procedure into an atomic skill set and to investigate new compositional reasoning chains, which ensure the model correctly invokes the necessary skill.

\subsection{Scientific Table-Claim Benchmarks}
Previous scientific fact-checking datasets have focused on text descriptions~\citep{wadden-etal-2020-fact,sarrouti-etal-2021-evidence-based,wang-etal-2023-check-covid}. Recently, there has been growing interest in scientific table or chart fact-checking, as a significant portion of critical information in scientific domain is conveyed through them~\cite{zhou-etal-2023-enhanced,huang-etal-2024-lvlms,pixel2insight_2025}. 
These papers use evidence from sources such as Wikipedia~\citep{gupta-etal-2020-infotabs,Chen2020TabFact}, news~\citep{akhtar-etal-2022-pubhealthtab}, and research papers~\citep{wang-etal-2021-semeval,lu-etal-2023-scitab}. However, these datasets usually ignore the reasoning processes involved in table fact-checking. In contrast, we annotate each claim-table pair with fine-grained atomic skills and long reasoning chains. Moreover, previous work focused on a single domain, including machine learning~\citep{lu-etal-2023-scitab} or biomedical domain~\citep{gupta-etal-2020-infotabs,Chen2020TabFact,wang-etal-2021-semeval,akhtar-etal-2022-pubhealthtab}, which limits their capacity to test the generalization ability of language models. 
To address this gap, we construct the first diverse datasets spanning multiple disciplines, including machine learning, materials science, medical science, and finance.

\section{SciAtomic Benchmark}
\begin{table}[t]
\small
    \centering
\begin{tabularx}{\linewidth}{XXXX}
\toprule
\textbf{Benchmark / Method} & \textbf{Domains} & \textbf{Training Samples}  & \textbf{Efficiency} \\
\midrule
Scigen & ML & / &  / \\
\hdashline
Sciab & ML & / &  / \\
\hdashline
FinQA & Fin & / & / \\
\hdashline
\textbf{SciAtomic} & ML, Med, Mat, Fin & 350  & \cmark \\
\hdashline
Tablellama & / & 2.6 m  & \xmark \\
\hdashline
Tablegpt & / & 2.4 m  & \xmark \\\hdashline
Tapex & / & 5 m  & \xmark \\\hdashline
TableLLM & / & 14 k & \xmark \\
\bottomrule
\end{tabularx}
    \caption{Comparison between SciAtomic and table claim verification benchmark and methods. ML denotes machine learning. Fin denotes finance. Mat denotes material science. Med denotes medical science. }
    \label{tab:my_label}
\end{table}
In this section, we formalize the task of scientific table-based claim verification and describe the construction of the \textsc{SciAtomic} benchmark.


\subsection{Problem Definition}

We study the task of \textit{scientific table-based claim verification}, where the goal is to determine whether a claim $C$ is \texttt{SUPPORTED} or \texttt{REFUTED} given a table $T$. Formally, a model $f_\theta(\cdot)$, parameterized by $\theta$, predicts a veracity label $Y = f_\theta(T, C)$, where $Y \in \{\texttt{SUPPORT}, \texttt{REFUTE}\}$. Each table $T = (P, \{T_{i,j}\})$ consists of a caption $P$ that provides domain-specific context, and cells $T_{i,j}$ organized in $R_T$ rows and $C_T$ columns. Claims are declarative scientific statements that may require quantitative reasoning, multi-step inference, or contextual interpretation from both the table and its caption.

Existing work has primarily focused on claims from computer science tables, limiting model robustness across scientific domains. To address this gap, we introduce the \textsc{SciAtomic} benchmark, which includes claims grounded in tables from underrepresented but high-stakes fields including materials science, medical science, and finance. All these domains pose diverse reasoning challenges and require precise, explainable verification to support scientific integrity and decision-making.

\subsection{Data Collection}

\noindent\textbf{Computer Science.} 
We construct the computer science subset of \textsc{SciAtomic} using the SciGen dataset~\cite{moosavi2021scigen}, which contains scientific tables and captions extracted from arXiv papers in the computer science domain. We sample 1,376 table-caption pairs, focusing on preserving diversity in structure and content.

\noindent\textbf{Finance.} 
We manually curate 343 tables from consolidated financial statements of randomly sampled S\&P 500 companies, including income statements, balance sheets, cash flows, and equity reports. Original data was captured via screenshots and converted to structured format using GPT-4-Vision.

\noindent\textbf{Medical Science.} 
We select 1,468 tables from PubTables-1M~\cite{smock2022pubtables} extracted from biomedical literature in PubMed Central Open Access, which ensures high cell-level fidelity through layout-aware parsing, making it well-suited for scientific claim grounding in the medical domain.

\noindent\textbf{Materials Science.} 
To represent the underexplored yet technically rich materials science domain, we incorporate 37 expert-annotated tables from MatSciTable~\cite{circi2024well}, mainly on polymer composites. Due to domain sparsity and depth, we generate multiple claims per table to capture diverse reasoning paths and emphasize domain-specific semantics for diverse and challenging verification. 

We present data statistics in Table~\ref{tab:data_statistics}. Further information on data collection is included in Appendix~\ref{app:collection_details}.

\begin{table}[b]
\vspace{-1em}
\small
\centering
\begin{tabularx}{\linewidth}{>{\hsize=1\hsize}X>{\arraybackslash\hsize=1\hsize}X>{\centering\arraybackslash\hsize=1\hsize}X>{\centering\arraybackslash\hsize=1\hsize}X>{\centering\arraybackslash\hsize=1\hsize}X>{\centering\arraybackslash\hsize=1\hsize}X
}
\toprule
 & \multirow{2}{*}{\textbf{SciTab}} & \multicolumn{4}{c}{\textbf{Our SciAtomic}} \\
\cmidrule(lr){3-6}
 &  & \textbf{ML.} & \textbf{Mat.} & \textbf{Med.} & \textbf{Fin.} \\
\midrule
\textbf{Neg.} & 411 & 400 & 160 & 364 & 381 \\
\textbf{Pos.} & 457 & 400 & 164 & 318 & 381 \\\hdashline
\textbf{Sum} & 868 & 800 & 324 & 682 & 762 \\
\bottomrule
\end{tabularx}
\vspace{-1em}
\caption{Statistics of Scitab and SciAtomic. \textit{Neg.} denotes negative claims. \textit{Pos.} denotes positive claims. \textit{ML.} denotes machine learning. \textit{Mat.} denotes material science. \textit{Med.} denotes medical science.  \textit{Fin.} is for finance. }
\label{tab:data_statistics}
\end{table}
\vspace{-0.5em}


\subsection{Human–Model Collaborative Annotation}
To efficiently generate high-quality claims while minimizing annotation burden, we adopt a human–model collaboration framework~\cite{lu-etal-2023-scitab}. Inspired by prior work, we use GPT-4o to produce both supported and refuted claims through structured prompting, followed by human validation. Please see Appendix~\ref{app:claim_annotation} for details.

\paragraph{Positive Claims.} We prompt GPT-4o to generate precise, multi-step scientific claims based on full tables, requiring reasoning over trends, derived metrics, and domain knowledge. Prompts enforce determinism and avoid vague expressions to ensure challenge and verifiability.

\paragraph{Negative Claims.} Refuted claims are created via minimal semantic flips and targeted data manipulations that subtly reverse meaning while preserving linguistic structure, simulating realistic scientific misinformation with the intents of concealing misinformation and amplifying harm. This strategy discourages shallow pattern matching and promotes robust model reasoning.

\begin{table*}[t]
\vspace{-2em}
\centering
\small
\begin{tabular}{llccccc}
\toprule
\textbf{Model} & \textbf{Size} & \textbf{SciTab} & \textbf{SciAtomic ML} & \textbf{Material} & \textbf{Medical} & \textbf{Finance} \\
\midrule
GPT-4o (Intuitive)
& 7b                & 0.4951          & 0.7513              & 0.6451    &0.6246 &0.5814\\
GPT-4o (CoT) 
& 7b                & 0.5507          & 0.9025              & 0.8580    &0.8152 &0.8570\\
\midrule
TAPEX 
&  4b              &  0.4060        &  0.5975             & 0.5264    & 0.5617 & 0.5197\\
\hdashline
TableLLaMa 
& 7b                & 0.5749         & 0.6263              & 0.5337    & 0.5710 & 0.5328\\
\hdashline

TableGPT2 
& 7b                & 0.6959         & 0.6750              & 0.5176    & 0.5432 & 0.5249\\
\hdashline
TableLLM (Text)
& 7b                &  0.4215        & 0.4913             & 0.4604    & 0.5000 & 0.4856\\
\midrule
\multirow{2}{*}{Phi-4}
  & 3.8b (base)     & 0.5184         & 0.5113         &0.5185              & 0.5166   & 0.5118 \\
  & 3.8b (ft)       & \cellcolor{newbleu}0.5472\textsuperscript{↑}         & \cellcolor{newbleu}0.7188\textsuperscript{↑}            & \cellcolor{newbleu}0.6356\textsuperscript{↑}              & \cellcolor{newbleu}0.5934\textsuperscript{↑}   & \cellcolor{newbleu}0.6181\textsuperscript{↑} \\
\hdashline
\multirow{4}{*}{LLaMA}
  & 3.2–3b (base)   & 0.5012         & 0.5038         & 0.4815        & 0.4956 & 0.4843 \\
  & 3.2–3b (ft)     & \cellcolor{newbleu}0.6106\textsuperscript{↑} & \cellcolor{newbleu}0.6688\textsuperscript{↑} & \cellcolor{newbleu}0.5740\textsuperscript{↑} & \cellcolor{newbleu}0.5176\textsuperscript{↑} & \cellcolor{newbleu}0.5276\textsuperscript{↑} \\
  & 3.1–8b (base)   & 0.4724         & 0.4925         & 0.4938        & 0.5059  & 0.4829 \\
  & 3.1–8b (ft)     & \cellcolor{newbleu}0.5910\textsuperscript{↑} & \cellcolor{newbleu}0.6513\textsuperscript{↑} & \cellcolor{newbleu}0.6296\textsuperscript{↑} & \cellcolor{newbleu}0.5381\textsuperscript{↑}  & \cellcolor{newbleu}0.5643\textsuperscript{↑} \\
\hdashline
\multirow{10}{*}{Qwen-2.5}
  & 1.5b (base)     & 0.5367         & 0.5675         & 0.4938       & 0.5308   &0.5210 \\
  & 1.5b (ft)       & \cellcolor{newbleu}0.5933\textsuperscript{↑} & \cellcolor{newbleu}0.6663\textsuperscript{↑} & \cellcolor{newbleu}0.5617\textsuperscript{↑} & \cellcolor{newbleu}0.5484\textsuperscript{↑}   & \cellcolor{newbleu}0.5866\textsuperscript{↑} \\
  & 3b (base)       & 0.5530         & 0.5163         & 0.4722         & 0.4765 &0.5827 \\
  & 3b (ft)         & \cellcolor{newbleu}0.5795\textsuperscript{↑}         & \cellcolor{newbleu}0.7000\textsuperscript{↑} & \cellcolor{newbleu}0.5988\textsuperscript{↑} & \cellcolor{newbleu}0.5308\textsuperscript{↑} & \cellcolor{newbleu}0.6247\textsuperscript{↑} \\
  & 7b (base)       & 0.4850        & 0.4975         & 0.4691        & 0.4985  & 0.4948 \\
  & 7b (ft)         & \cellcolor{newbleu}0.5956 \textsuperscript{↑}        & \cellcolor{newbleu}0.7100\textsuperscript{↑} & \cellcolor{newbleu}0.6821\textsuperscript{↑} & \cellcolor{newbleu}0.6202\textsuperscript{↑}  & \cellcolor{newbleu}0.6562\textsuperscript{↑} \\
  & 14b (base)      & 0.6578        & 0.7550         & 0.6296            & 0.6085  & 0.5932 \\
  & 14b (ft)        & \cellcolor{newbleu}\textbf{0.7009}\textsuperscript{↑}            & \cellcolor{newbleu}0.8025\textsuperscript{↑}         & \cellcolor{newbleu}0.7130\textsuperscript{↑}             & \cellcolor{newbleu}0.7067\textsuperscript{↑}  & \cellcolor{newbleu}0.7165\textsuperscript{↑} \\
\midrule
\multirow{2}{*}{Deepseek-R1-LLaMA}
  & 8b (base)       & 0.5611         & 0.6663         & 0.6049             & 0.5689  & 0.5958 \\
  & 8b (ft)         & \cellcolor{newbleu}0.6129\textsuperscript{↑}            & \cellcolor{newbleu}0.7150\textsuperscript{↑}         & \cellcolor{newbleu}0.6512\textsuperscript{↑}             & \cellcolor{newbleu}0.6276\textsuperscript{↑}  & \cellcolor{newbleu}0.6430\textsuperscript{↑} \\
\hdashline
\multirow{4}{*}{Deepseek-R1-Qwen}
  & 7b (base)       & 0.5853         & 0.6300         & 0.5895              & 0.5411   & 0.6312 \\
  & 7b (ft)         & \cellcolor{newbleu}0.5924\textsuperscript{↑} & \cellcolor{newbleu}0.8063\textsuperscript{↑} & \cellcolor{newbleu}0.7593\textsuperscript{↑}           & \cellcolor{newbleu}0.7331\textsuperscript{↑}   & \cellcolor{newbleu}\textbf{0.8570}\textsuperscript{↑} \\
  & 14b (base)      & 0.6560         & 0.7500         & 0.6728              & 0.6818   & 0.7205 \\
  & 14b (ft)        & \cellcolor{newbleu}0.6613\textsuperscript{↑}         & \cellcolor{newbleu}\textbf{0.8200\textsuperscript{↑}} & \cellcolor{newbleu}\textbf{0.7653}\textsuperscript{↑}              & \cellcolor{newbleu}\textbf{0.7654}\textsuperscript{↑}   & \cellcolor{newbleu}0.8045\textsuperscript{↑} \\
\bottomrule
\end{tabular}
\caption{Performance comparison between LLMs fine-tuned on 350 training samples and their base versions, and SOTA LLMs. 
↑ indicates gain after fine-tuning. The evaluation metric is prediction accuracy.}
\label{tab:llm-perf}
\vspace{-1em}
\end{table*}


\section{Atomic Reasoning for Claim Verification}

In this section, we introduce \textbf{AtomicTable}, a skill-chaining generation schema tailored for table-based scientific claim verification. AtomicTable is designed to improve the reasoning capability of LLMs through a fine-grained and modular decomposition of the verification task. Our motivation stems from two complementary dimensions: the nature of the task itself and the way LLMs process complex reasoning.

From the \textit{task decomposition} perspective, table-based claim verification typically involves multiple distinct and interleaved reasoning operations, such as matching entities, aggregating values, interpreting structures, or identifying causal patterns. Each of these operations can be framed as a discrete subtask. By explicitly dividing the overall verification into such well-scoped subtasks, we enable more interpretable and controllable reasoning, aligning with the divide-and-conquer principle in human cognitive strategies.

From the \textit{model capability} perspective, we draw inspiration from Cognitive Load Theory, which emphasizes reducing the burden on working memory during complex tasks. We achieve this by defining a set of \textit{atomic reasoning skills}-modular, reusable units of reasoning that model can invoke when needed. Instead of tackling the entire claim holistically, the model proceeds step-by-step, applying only the relevant skills at each stage. This modularization helps constrain the search space, reduce overfitting to spurious patterns, and improve generalization to unseen claims and table structures.

\subsection{Skill-Chain Schema}
\label{sectable:chain}

We formalize our skill-chain schema as:
\vspace{-0.5em}
\begin{small}
\begin{align*}
\text{interpretation} \rightarrow \text{planning} \rightarrow 
[\text{subplan} \rightarrow \text{cell grounding} \\
\rightarrow \text{reasoning} \rightarrow \text{recap}]^N \rightarrow \text{conclusion}
\end{align*}
\end{small}
In the schema, the model first interprets the overall verification goal and generates a high-level plan with $N$ subgoals. Each subgoal is processed through a local reasoning loop involving evidence grounding, atomic reasoning, and result summarization. The outcomes of all subgoals are then synthesized into a final verdict.

\noindent $\bullet$ \textbf{Interpretation:} Given table $T$ and claim $C$, model produces interpretation of verification task:

\vspace{-0.5em}
\begin{small}
\[
I = f^\text{interp}_\theta(T, C)
\]
\end{small}

\noindent $\bullet$ \textbf{Planning:} Based on the interpretation, the model generates a set of $N$ subplans or subgoals:

\begin{small}
\[
P = \{p_1, p_2, \ldots, p_N\} = f^\text{plan}_\theta(T, C, I)
\]
\end{small}

\noindent $\bullet$ \textbf{Cell Grounding:} For each subplan $p_i$, the model identifies the relevant cells in the table as evidence:

\vspace{-0.5em}
\begin{small}
\[
G_i = f^\text{ground}_\theta(T, C, p_i)
\]
\end{small}

\noindent $\bullet$ \textbf{Reasoning:} The model applies appropriate atomic reasoning skills to the grounded evidence:

\begin{small}
\[
R_i = f^\text{reason}_\theta(T, C, p_i, G_i)
\]
\end{small}

\noindent $\bullet$ \textbf{Recap:} The result of the reasoning is summarized as a local outcome:

\vspace{-0.5em}
\begin{small}
\[
U_i = f^\text{recap}_\theta(T, C, \{R_j\}_{j \leq i}, p_i)
\]
\end{small}

\noindent $\bullet$ \textbf{Conclusion:} Finally, all local recaps are aggregated into a global decision:

\vspace{-1em}
\begin{small}
\[
Y = f^\text{final}_\theta(T, C, \{U_i\}_{i=1}^N), \quad Y \in \{\text{SUPPORT}, \text{REFUTE}\}
\]
\end{small}
\vspace{-2em}

This structured reasoning flow offers three key benefits: (1) each step focuses only on its local context, reducing distraction from irrelevant history; (2) the use of atomic skills avoids unintended entanglement of reasoning functions; (3) the modular chain improves interpretability and robustness.

\subsection{Atomic Skill Set}
\label{sectable:skill}

Each reasoning step in our schema is realized as a composition of atomic reasoning skills, enabling the model to perform targeted inference based on local task demands:

\vspace{-0.5em}
\begin{small}
\[
R_i = f^\text{reason}_{\theta}\bigl(\{s_k\}_{k \in A_i};\, T, C, p_i, G_i\bigr)
\]
\end{small}
\vspace{-1em}

\noindent \(A_i \subseteq \{1, \ldots, K\}\) denotes the index set of atomic skills selected at step \(i\); each \(s_k\) represents a distinct reasoning module tailored to a specific inference type.
These skills serve as the foundation for modular and interpretable reasoning. By isolating key capabilities into lightweight, reusable components, we allow model to invoke only the necessary skills at each step, reducing cognitive burden and improving both precision and generalization. This design also enhances transparency, as the reasoning trace can be clearly attributed to distinct, well-defined competencies.
We define following atomic skill set, capturing the core operations commonly required for table-based scientific claim verification:

\noindent $\bullet$\textbf{Conceptual Understanding:} Interpreting domain-specific language and aligning abstract claims with table semantics.

\noindent $\bullet$ \textbf{Structure Analysis:} Analyzing the organization of the table, including row-column relationships and hierarchical headers.

\noindent $\bullet$ \textbf{Numerical Analysis:} Performing quantitative operations such as comparisons, aggregations, and unit normalization.

\noindent $\bullet$ \textbf{Causal Analysis:} Inferring causal or correlational relations implied by data patterns and trends.

This atomic design not only supports compositional reasoning but also improves the system's adaptability across diverse table types and scientific contexts. By breaking complex inference into smaller, skill-specific steps, our framework fosters more accurate, robust, and explainable verification.


\subsection{Skill-Chain Evaluation Principle}
Building on our skill-chain formulation, we introduce a comprehensive evaluation framework comprising five dimensions to assess the quality of model-generated reasoning chains: granularity, information redundancy, alignment, interpretability, and accuracy. Detailed definitions are in Appendix~\ref{ssec:evalPrinciple}.

\noindent \textit{\textbf{Granularity}} measures the fineness of each reasoning step. Fine-grained reasoning enables precise extraction of knowledge from dense tabular content and supports higher modularity in skill application.

\noindent \textit{\textbf{Information Redundancy}} quantifies the presence of superfluous or irrelevant information within the reasoning chain. Minimizing redundancy is critical to improving inference efficiency and scalability in real-world scientific applications.

\noindent \textit{\textbf{Alignment} }captures the logical coherence between adjacent reasoning steps. Strong alignment ensures logical progression between steps, promoting consistency and reducing reasoning drift.

\noindent \textit{\textbf{Interpretability}} reflects how understandable the reasoning process is to human readers, especially non-experts. In scientific domains, tables have high information density, and clear reasoning chains are essential for transparency and trustworthiness.

\noindent \textit{\textbf{Accuracy}} evaluates the correctness of each individual step within the reasoning chain, providing a fine-grained perspective on reasoning fidelity.

\begin{figure}[t]
\vspace{-0.5em}
    \centering    \includegraphics[width=0.49\linewidth]{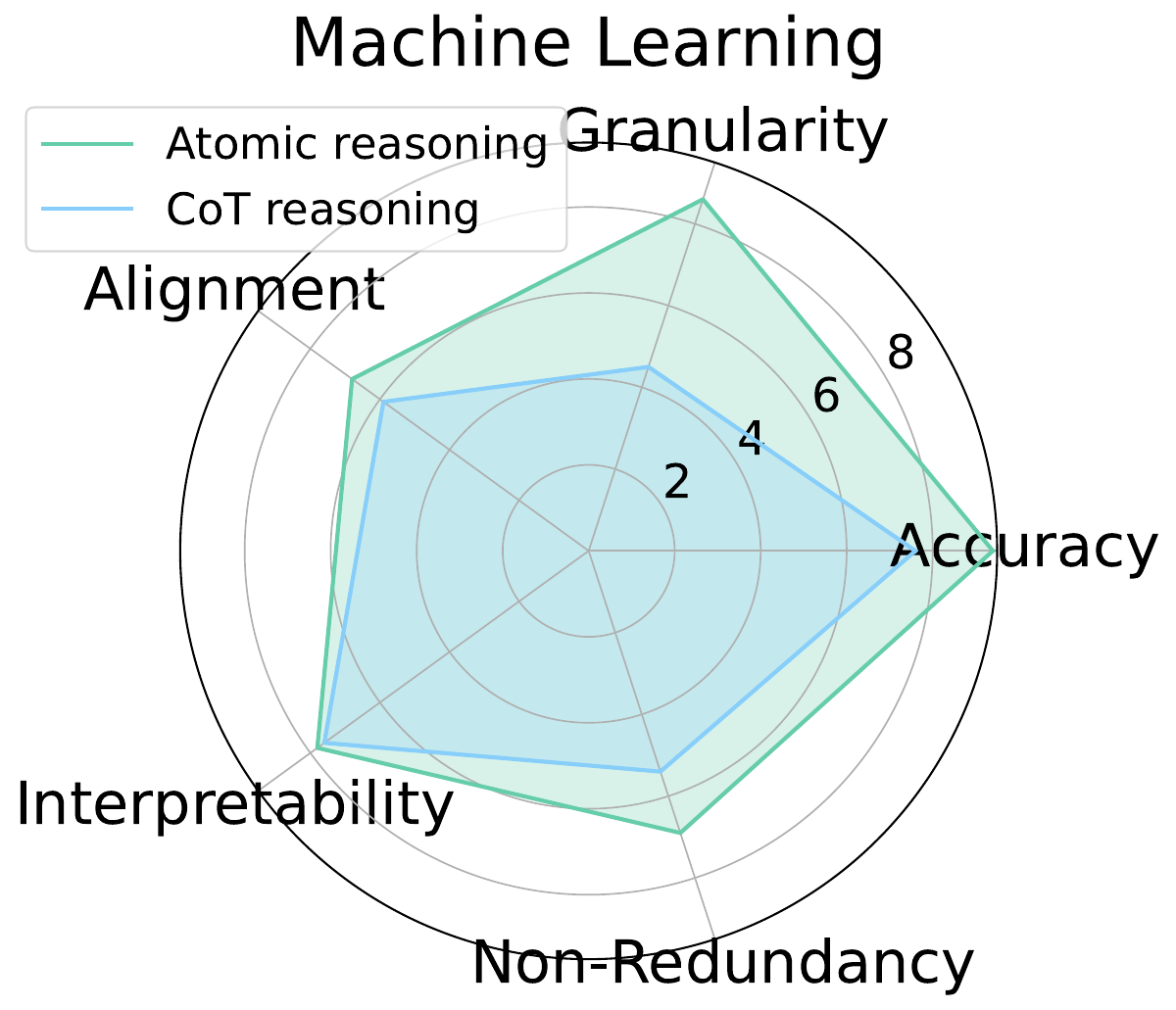}
    \includegraphics[width=0.49\linewidth]{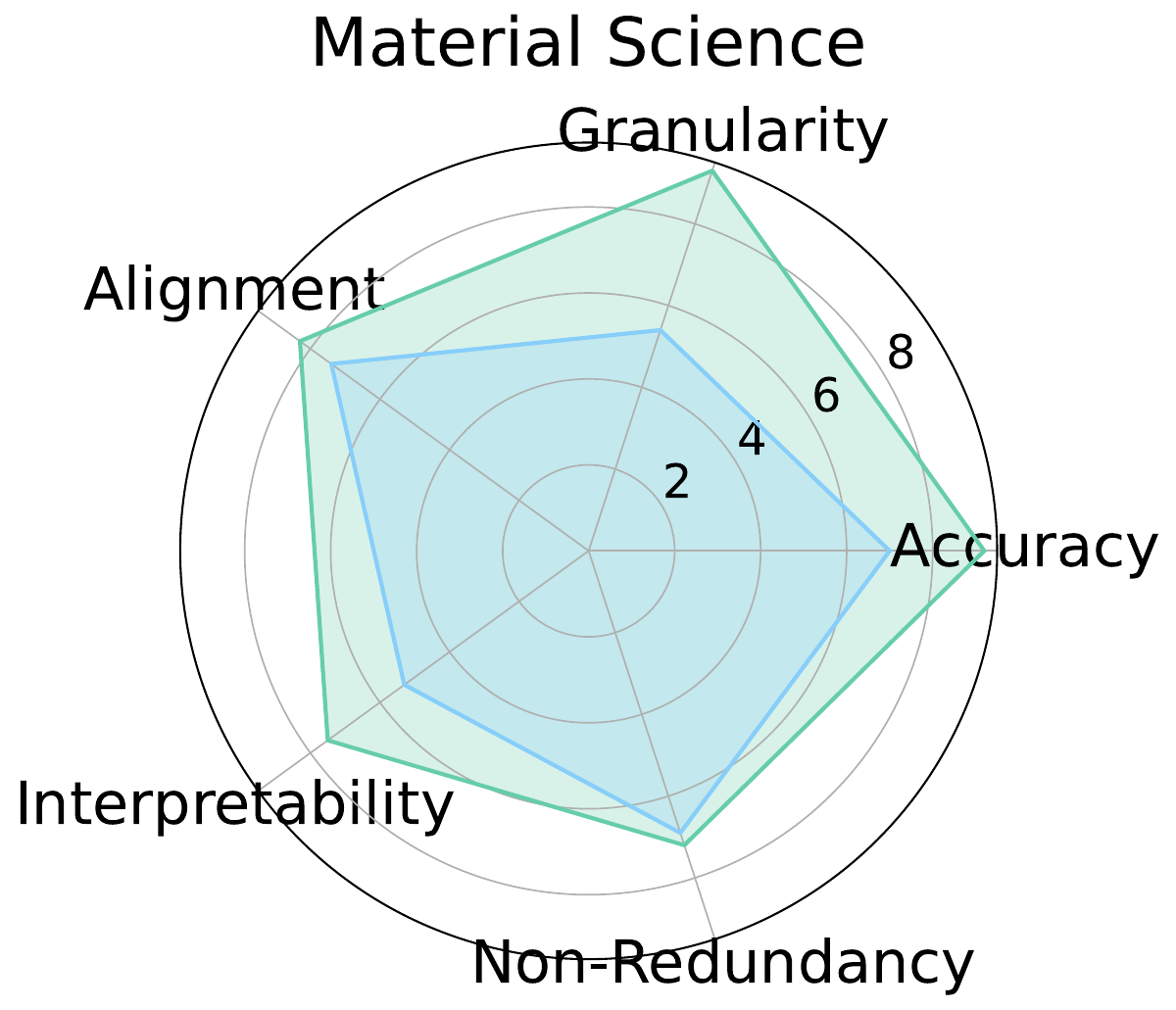}
\vspace{1em}
\includegraphics[width=0.49\linewidth]{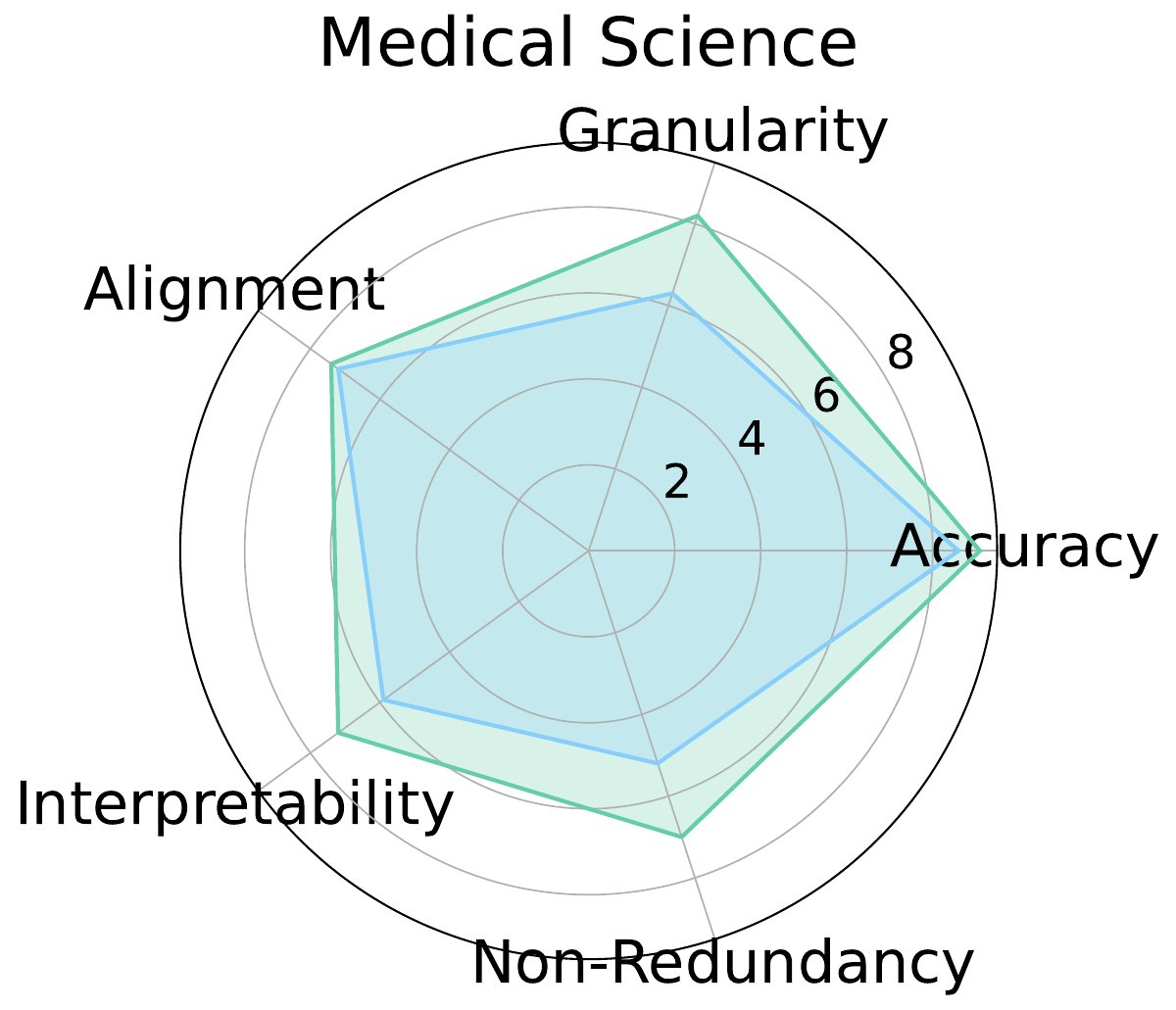}
    \includegraphics[width=0.49\linewidth]{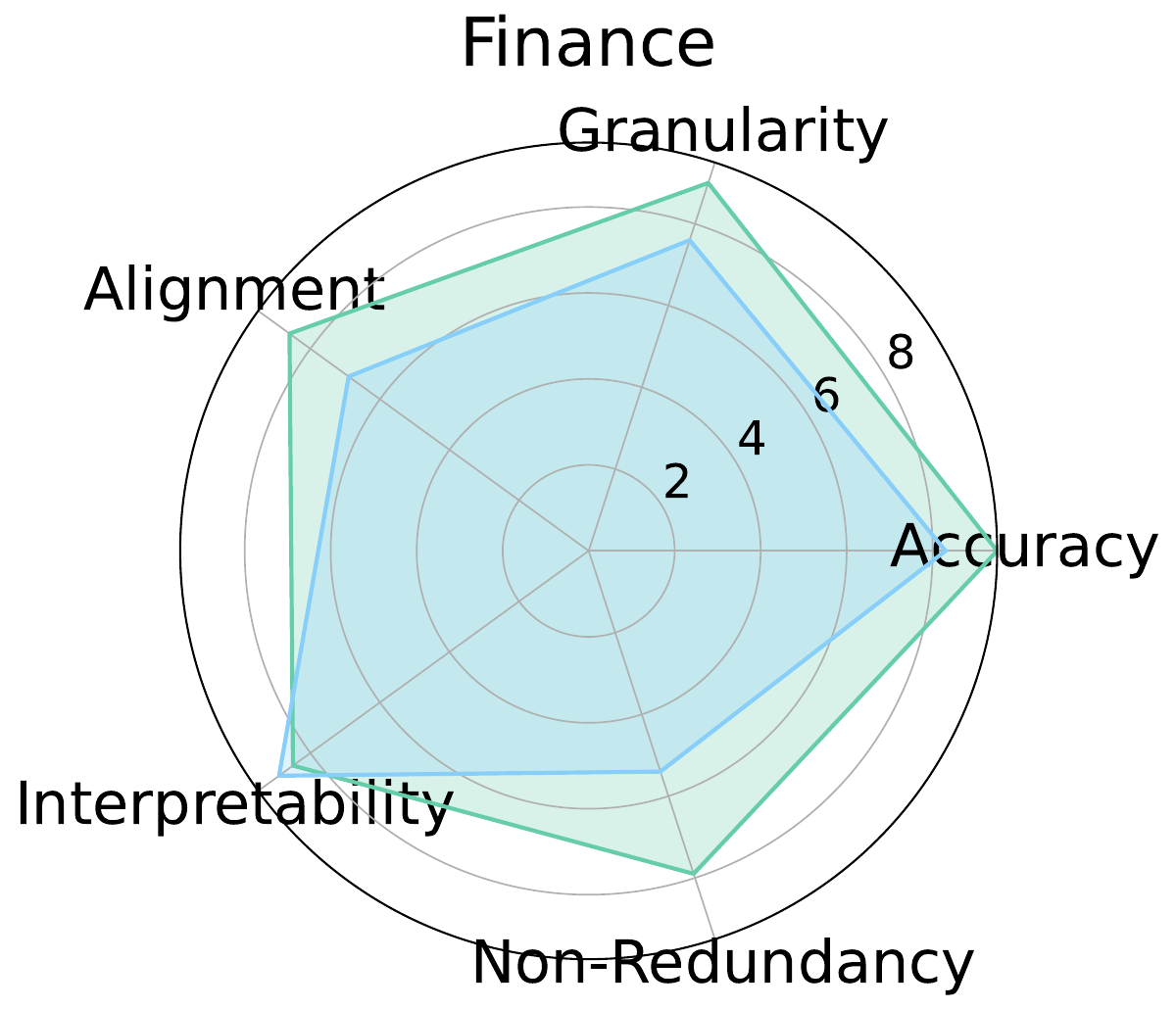}
    \vspace{-2em}
\caption{Quality evaluation for GPT-4o CoT reasoning and our atomic reasoning chain over five dimensions.}
\label{fig:qualityRadar}
\vspace{-1.3em}
\end{figure}

The comparison evaluation results of our atomic reasoning chain and GPT-4o chain-of-thoughts are shown in Figure~\ref{fig:qualityRadar}. The accuracy, non-redundancy, and alignment are evaluated by GPT-4o over three runs and averaged. Granularity and interpretability are assessed by three human annotators on a 0–10 scale, and their scores are averaged.

\subsection{Atomic Skill Distribution Analysis}

\begin{figure}[b]
    \vspace{-0.5em}\centering    \includegraphics[width=1.0\linewidth]{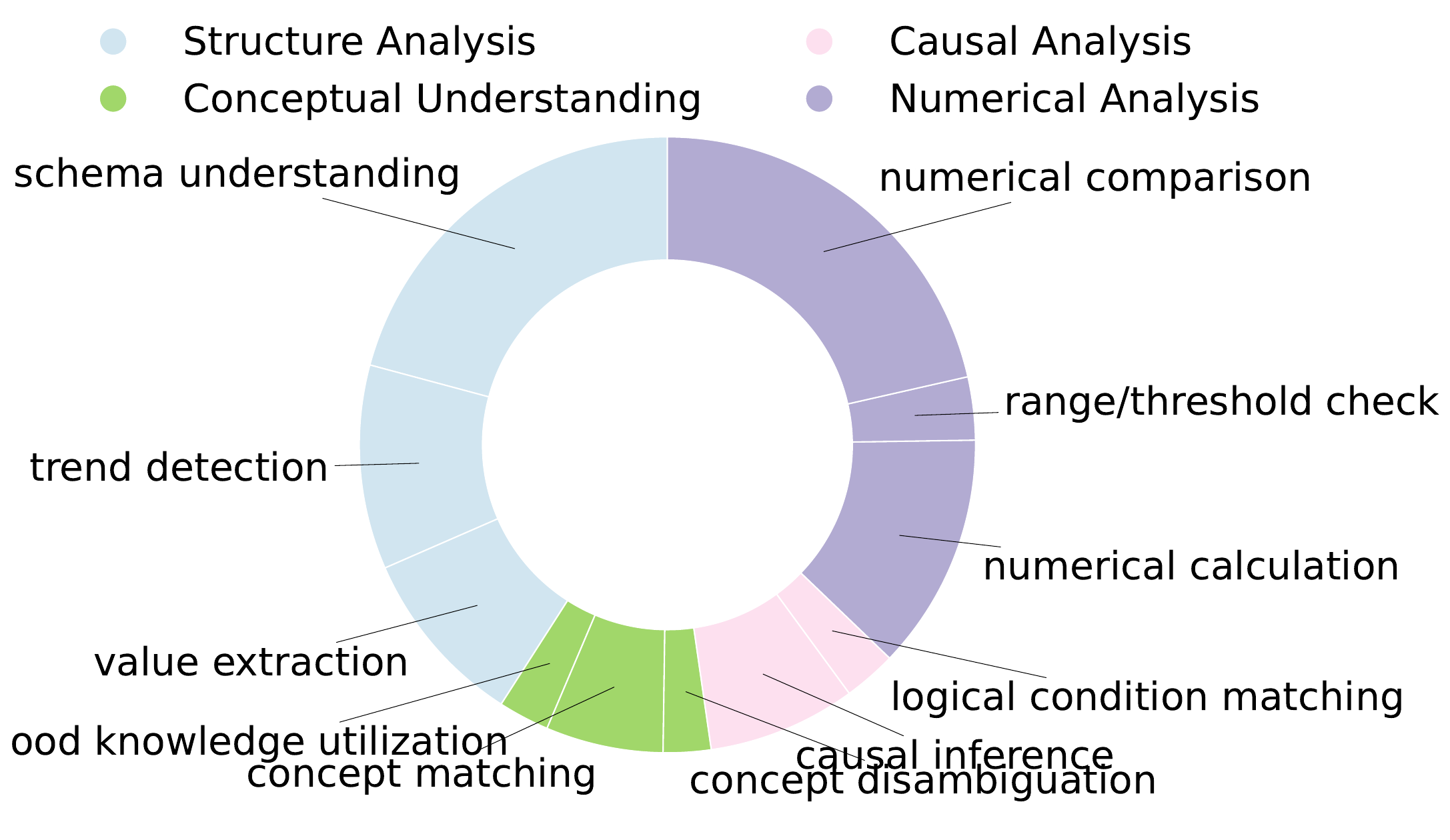}
\vspace{-1.8em}
\caption{Atomic skill distribution in Material Science.}
\label{fig:matPie}
\end{figure}

We analyze atomic skill distribution in Figure~\ref{fig:matPie}. Despite the domain-specific complexity, most claims are resolved using a small, consistent set of numerical and structural skills. This suggests complex reasoning can be decomposed into compact, reusable units with strong generalizability.
\section{Experiment}

\subsection{Comparison Baselines}
We compare our model with \textbf{(1) state-of-the-art Closed-Source LLM}, including GPT-4 \cite{openai2023gpt4}, then compare our model with the \textbf{(2) Table-specialized LLMs}, including TAPEX~\cite{liu2022tapex}, TableLlama~\cite{zhang-etal-2024-tablellama}, TableGPT~\cite{su2024tablegpt2largemultimodalmodel} and TableLLM~\cite{zhang2024tablellm}. We also include \textbf{(3) various LLMs for ablation}, including Phi-4~\cite{abdin2024phi}, Qwen-2.5~\cite{qwen2024qwen2}, LLaMA~\cite{grattafiori2024llama}, and reasoning-enhanced Deepseek-R1 series models~\cite{guo2025deepseek}.








\subsection{Implementation Details}
We finetuned the large language models (LLMs) using 350 training samples and 50 validation samples from the machine learning domain. The models were trained over 3 epochs with a learning rate of 1e-5. For text generation, we set the temperature to 0.8 and applied top-k sampling with k=0.9.


\subsection{Main Results}
We evaluate the effectiveness of atomic reasoning by comparing base LLMs to their fine-tuned counterparts using our atomic reasoning chains. We benchmark our atomic LLMs against existing state-of-the-art models on the SciAtomic dataset.

\paragraph{Results of Fine-tuning LLMs.}
In Table~\ref{tab:llm-perf}, our key observations are as follows:

\textit{(1) Atomic reasoning improves all base LLMs, even with only 350 training examples.}
This highlights the efficiency of atomic supervision, which yields strong performance gains with limited data.

\textit{(2) Even reasoning-specialized LLMs benefit substantially from atomic supervision.}
The improvements persist across models of different sizes and pretraining objectives, indicating that atomic supervision offers complementary inductive biases beyond standard chain-of-thought prompting.

\textit{(3) Atomic reasoning equips LLMs with strong 
cross-domain generalization capabilities.}
On out-of-domain evaluation sets, fine-tuned models consistently outperform their base counterparts, demonstrating improved adaptability.

\paragraph{Results of SOTA Baselines.}
Beyond evaluating the performance gains introduced by atomic reasoning, we also assess how existing state-of-the-art models perform on our SciAtomic benchmark.

\textit{(4) Existing state-of-the-art table-claim verification models struggle on our SciAtomic benchmark.} Despite being trained on large-scale fact verification datasets, these models exhibit clear performance gaps when faced with the fine-grained, compositional reasoning required by SciAtomic. In contrast, our atomic fine-tuned models surpass TableLlama while using \textbf{7,429} times fewer training samples.
This underscores the unique challenge posed by our benchmark and the need for more interpretable and faithful reasoning supervision.

\subsection{Atomic Reasoning in Action: Emergence, Scaling, and Failure Modes}

In this section, we further analyze atomic reasoning on scientific table claim verification, including unseen skills emerging in new domains, efficient training and inference scaling, and error analysis across CoT and atomic reasoning.

\begin{figure}[htb]
    \centering
    \vspace{-0.5em}\includegraphics[width=0.98\linewidth]{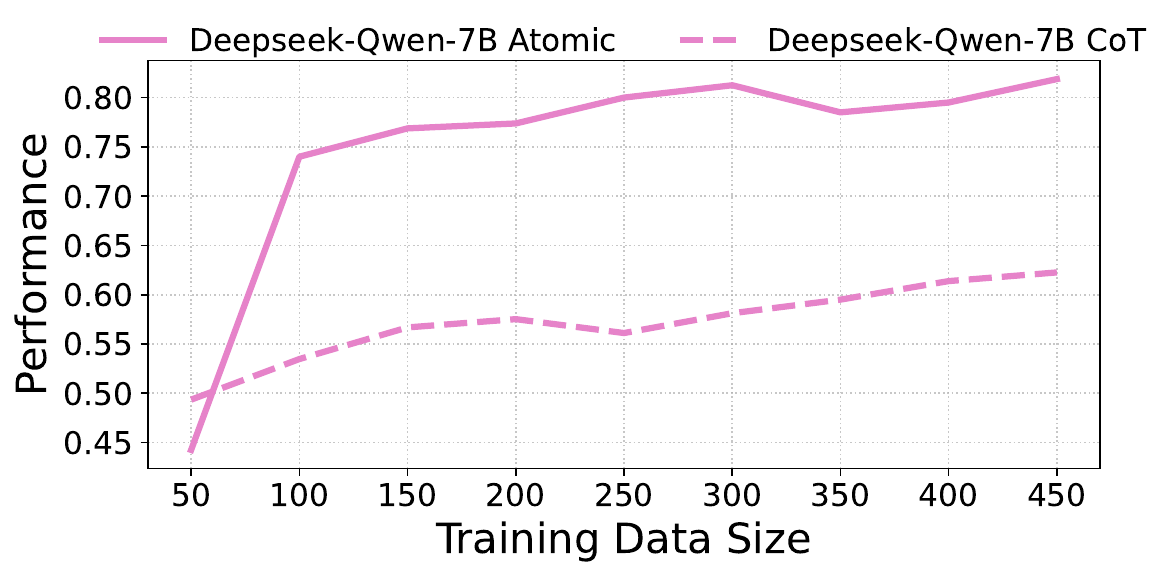}
    \includegraphics[width=1.04\linewidth]{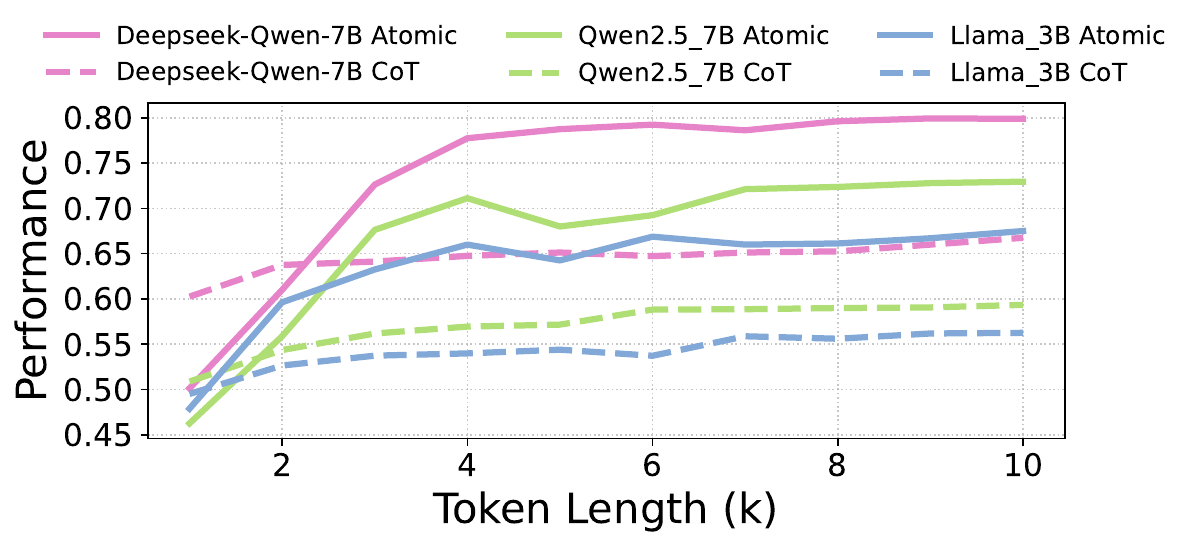}
    \vspace{-1.8em}
    \caption{Training (upper) and Inference (lower) time scaling effect on the Machine Learning domain.}
\label{fig:scaling}
\vspace{-1.3em}
\end{figure}
\paragraph{Emerging Skills.}

Training models on reasoning chains with fixed atomic skills enhances cross-domain generalization in our case study. Remarkably, during inference, models not only apply known skills but also exhibit emergent behavior by composing novel or compound skills. For instance, after training DeepSeek-R1-Qwen-7B on 350 atomic reasoning chains from the machine learning domain, the model demonstrates emergent skill composition when evaluated on material science.
Given a table on the surface properties of PI/OFG Nanocomposites, the model's generated reasoning chain reveals that it has correctly identified the need to compute the rate of change in y's value as OFG increases. Interestingly, it also incorporates its understanding of table structure and the adjacency of OFG feed ratios to execute a compound skill. This skill combines value extraction, numerical calculation, and schema understanding to produce a reasoning step such as: ``Compute differences in y's ${(mN/m)}^2$ between \textbf{adjacent} OFG feed ratios... the average decrease is –1.2\%.''


\paragraph{Training-time Scaling.}
As shown in Figure~\ref{fig:scaling}, our skill-chain schema achieves higher training efficiency than CoT by guiding the model to use modular, reusable atomic skills. This structured approach enables better generalization with fewer examples and more scalable supervision.

\paragraph{Inference-time Scaling}

Our skill-chain schema is explicitly designed to enforce logical progression between reasoning steps, which helps avoid redundant attention on irrelevant history and prevents unintended skill activations. This localized dependency improves inference efficiency. As shown in Figure~\ref{fig:scaling}, LLMs on atomic reasoning scale more efficiently than CoT reasoning in inference time.

\paragraph{Error Analysis. }
We conduct error analysis comparing CoT and atomic reasoning across three categories: snowball errors, contextual conflicting errors, and coarse-grained errors.

\begin{figure}[htb]
    \centering
\vspace{-1em}    \includegraphics[width=1.05\linewidth]{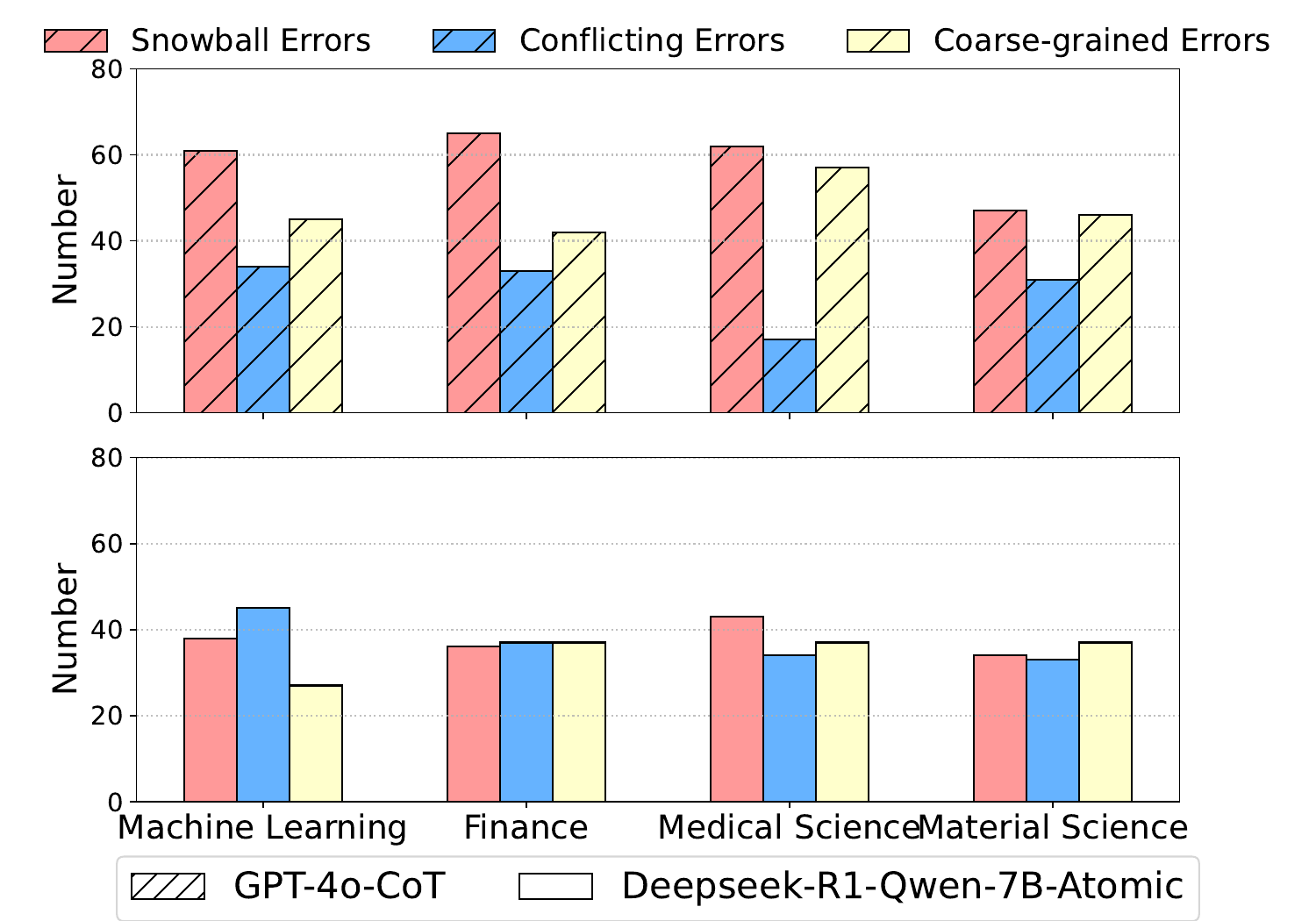}
    \vspace{-1.8em}
    \caption{Human evaluation of three reasoning error types on 100 failed reasoning samples, respectively by DeepSeek-R1-Qwen-7B-Atomic and GPT-4o-CoT.}
\label{fig:error}
\vspace{-1.3em}
\end{figure}

\textit{Snowball errors: }
Figure~\ref{fig:error} shows that Deepseek-R1 with atomic reasoning makes fewer snowball errors than GPT-4o's CoT. This is due to the skill-chain schema's localized dependencies and modular skill use, which limit error propagation.

\textit{Conflicting errors: }
Contextual conflict arising from contradictions between contextual information and inferred facts occurs at a comparable rate in both CoT and atomic reasoning, suggesting that our method maintains consistency without sacrificing factual alignment.

\textit{Coarse-grained errors: }
Coarse-grained errors caused by reasoning that skips intermediate steps or aggregates multiple operations are significantly reduced in our atomic reasoning approach, benefiting from its fine-grained and step-wise structure.

\section{Conclusion}
We present atomic reasoning, a data- and scaling-efficient paradigm that enables language models to learn modular, composable reasoning skills for scientific table-claim verification. By decomposing complex verification into modular atomic skills and introducing a skill-chaining schema, our approach improves generalization across claim types and scientific domains. We also release SciAtomicBench, a cross-domain benchmark with fine-grained reasoning annotations, facilitating rigorous evaluation. Our table-specialized model, AtomicTableLLM, achieves state-of-the-art performance, demonstrating the effectiveness of atomic reasoning in enhancing both reasoning accuracy and data efficiency.

\section{Limitations}
\label{sec:limitation}

We discuss various limitations throughout the paper. Here, we provide additional details. Our data was collected from open-access datasets written in English, with the number of available instances constrained by the original sources. In the future, we aim to extend our models to include table-claim pairs in other languages and domains. Due to changes in the API, the inherent randomness of the GPT-4 model and human annotation, our annotated dataset may not be easily reproducible, so we will release our dataset. Furthermore, limited computational resources prevented us from conducting experiments on larger models. Finally, our annotators were recruited from Ph.D. students, whose perspectives may differ from those of individuals with other levels of domain expertise.

\section*{Ethical Considerations}
In this paper, we present a method for verifying claims based on scientific tables, ensuring factual accuracy in scientific communication. Our approach achieves effective claim verification using a limited training set of just 350 instances, without relying on extensive instruction-tuning annotations. By minimizing dependence on manual labeling, our method promotes fairness, scalability, and inclusivity in AI for Science, contributing to the broader democratization of large language models (LLMs) across global communities.  Despite our method's attempts to reduce misinformation, our generation results may still suffer from hallucinations.

\bibliography{custom,anthology}

\appendix


\section{Benchmark Collection Details}
\label{app:collection_details}

\paragraph{Computer Science. } 
We constructed the computer science subset of our SciAtomic benchmark using the publicly available SciGen dataset~\cite{moosavi2021scigen}, which consists of scientific tables and associated captions extracted from arXiv papers in the computer science domain.
We sampled 1,376 table-caption pairs from the SciGen dataset (we discuss data quality control in more detail in \S\ref{ssec:data_quality_control}). 
 
\paragraph{Finance.} 
For the finance domain, we manually curated tables from consolidated income statements, balance sheets, statements of shareholders' equity, and cash flow statements of randomly sampled S\&P 500 companies. The data was originally captured as screenshots and converted to structured format using GPT-4-Vision. Unlike prior financial datasets that have focused on small or simplified tables, our collection reflects the full complexity of complete corporate financial reports. In total, 343 high-fidelity tables were extracted and standardized into a unified format.

\paragraph{Medical Science.} We selected 1,468 medical tables from PubTables-1M~\cite{smock2022pubtables} to construct a medical table-claim dataset. These tables were originally sourced from scientific papers in PubMed Central Open Access, with PubTables' extraction system ensuring high cell-level accuracy.

\paragraph{Material Science.} 
Tables in the materials science domain encode rich information such as material properties and composition, which poses unique challenges for scientific table-claim verification. However, this domain remains underrepresented in existing benchmarks, largely due to the difficulty of obtaining clean, structured tabular data. Automated extraction methods often struggle with cell-level accuracy, particularly in capturing numerical values, measurement units, and experimental descriptors. Hence, we incorporate 37 high-quality, expert-annotated tables released by MatSciTable~\cite{circi2024well}, which focus on polymer composites. To mitigate data sparsity and fully exploit the rich content of each table, we construct multiple claims per table, reflecting diverse reasoning trajectories grounded in domain-specific semantics.

\section{Claim Annotation Details}
\label{app:claim_annotation}

\subsection{Positive Claim Generation}

We aim to construct challenging, multi-step claims that reflect real scientific reasoning. For each table, we use a structured prompting framework with ChatGPT~\cite{openai2023gpt4}, incorporating a multi-stage instruction. The prompt instructs the model to:
\begin{itemize}
    \item Identify non-trivial patterns (e.g., extrema, trends, ratios).
    \item Integrate domain knowledge and contextual cues from the caption.
    \item Formulate declarative claims requiring at least five implicit reasoning steps.
\end{itemize}
To ensure clarity and verifiability, we explicitly discourage vague or subjective language (e.g., ``significantly better'') and instead encourage precise comparisons (e.g., ``12.3\% higher than''). Generated claims omit intermediate calculations and direct cell references to mirror authentic scientific discourse. Each generated claim undergoes human review for factual correctness, clarity, and linguistic naturalness.

\subsection{Negative Claim Generation}

To construct refuted claims that simulate real-world scientific misinformation, we implement two complementary strategies:

\paragraph{Minimal Semantic Flip.} Instead of generating false claims from scratch, which risks introducing lexical artifacts, we prompt ChatGPT to minimally edit true claims in a way that reverses their meaning. This retains the original syntax and structure, making errors subtle and harder to detect.

\paragraph{Targeted Data Manipulation.} We also generate refuted claims by altering key quantitative elements (e.g., thresholds, magnitudes, units) in a way that contradicts the underlying table data while preserving the claim's form. This approach simulates harmful misreporting (e.g., flipping safety limits or outcome labels), requiring models to perform deep reasoning over the table to detect inconsistencies.

Together, these strategies produce a challenging and realistic set of negative claims that promote robustness in verification models.

\subsection{Data Quality Control}
\label{ssec:data_quality_control}

In scientific domains, rigor and contextual specificity are critical. For example, when comparing material performance, a single table may report multiple metrics under varying experimental conditions, where omitting such context can introduce ambiguity and misinterpretation. Although existing scientific table claim verification benchmarks~\cite{wang-etal-2021-semeval,lu-etal-2023-scitab} have laid important groundwork, they place less emphasis on the verification of claims that require fine-grained, context-sensitive reasoning. In contrast, our benchmark prioritizes the construction and validation of scientifically rigorous claims, with careful attention to eliminating ambiguity and preserving essential domain-specific conditions.

\paragraph{Resolve Ambiguity and Out-of-Scope Issues.}

To resolve ambiguity, we performed multiple rounds of claim rewriting, implicit reference rewriting, and contextual information addition. We also implemented an out-of-scope (OOS) removal process to revise OOS claims while maintaining verifiability. We first identified claims requiring external scientific context beyond the table data, often containing implicit references or domain-specific terminology. For each identified OOS claim, ChatGPT-4~\cite{openai2023gpt4} then performed rewrites to eliminate dependencies on external knowledge while preserving the claim's core meaning and verifiability against the table.




\paragraph{Cross-Validation.}
We used multiple-path consistency to cross-validate long-thought, short-thought, and human-thought processes, ensuring high data quality.
If inconsistency occurs, another annotator will summarize existing procedures and give a final label.

\subsection{Data Evaluation Principle}
\label{ssec:evalPrinciple}

\begin{itemize}
    \item \textit{Accuracy:} We define the step-wise accuracy as the average correctness of all reasoning steps. Each step $f^\text{step}_\theta$ is considered accurate if its output $R_i$ is judged correct by SOTA LLMs. Formally,
\[
\text{Accuracy} = \frac{1}{N} \sum_{i=1}^{N} \mathbb{I}\left[\text{ModelCheck}(R_i)\right]
\]
where $\text{ModelCheck}(R_i)$ denotes that the output of step $i$ is labeled as correct by SOTA LLMs, and $N$ is the number of steps.
    \item \textit{Granularity:} Granularity characterizes the minimal yet sufficient scope of context $\mathcal{C}_i$ needed to support accurate execution of a reasoning step, which is scored by human annotators with the score range as 0-10.
This trade-off ensures that each reasoning unit operates over compact, focused input while preserving correctness, enabling modular decomposition without loss of fidelity.
    \item \textit{Interpretability:} We define interpretability as the degree to which each reasoning step $f_\text{step}$ can be semantically and logically followed by a human reader. A reasoning step is considered interpretable if a human, given the same local context $\mathcal{C}_i$, can reproduce or validate the step. Formally, we define:
\[
\text{Interp}_i = \mathbb{I}\left[ f_\text{step}(T, C, \mathcal{C}_i) = \hat{R}_i^{\text{human}} \right]
\]
where $\hat{R}_i^{\text{human}}$ denotes the reasoning outcome produced independently by human annotators based solely on the input context $\mathcal{C}_i$. High interpretability ensures that the model's reasoning chain can be audited and trusted.
    \item \textit{Information Redundancy:} We define information redundancy as the presence of non-essential elements in the local context $\mathcal{C}_i$ that do not contribute to reasoning at step $i$. Specifically, an input $x \in \mathcal{C}_i$ is redundant if:
\[
f^\cdot_\theta(T, C, \mathcal{C}_i) = f^\cdot_\theta(T, C, \mathcal{C}_i \setminus \{x\})
\]
Our data construction explicitly eliminates such redundancy by enforcing that every input token is necessary for the model's output:
\[
\forall x \in \mathcal{C}_i, \quad f^\cdot_\theta(T, C, \mathcal{C}_i) \neq f^\cdot_\theta(T, C, \mathcal{C}_i \setminus \{x\})
\]

    \item \textit{Adequate Alignment:} Adequate alignment ensures that each reasoning step $f_\text{step}$ correctly incorporates and builds upon the information from previous steps, maintaining the logical consistency throughout the reasoning process. Formally, we define the alignment at each step as follows:
For a given reasoning step $i$, the reasoning outcome $R_i$ should align with the accumulated knowledge up to step $i-1$ and should not introduce any contradictions or errors in the context. Specifically:
\[
\text{Align}_i = \mathbb{I}\left[ f_\text{step}(T, C, \mathcal{C}_i) = f_\text{step}(T, C, \mathcal{C}_{i-1}) \right] 
\]
where $\mathcal{C}_i$ represents the context and information accumulated until step $i$, and $f_\text{step}$ denotes the reasoning step at the $i$-th stage. High adequate alignment means that the model's reasoning steps are consistent with prior steps and no relevant information is lost or incorrectly altered during the process. If $\text{Align}_i = 1$, the reasoning step is aligned correctly; otherwise, it is misaligned.

\end{itemize}

\section{Annotation Details}

We invited four volunteers with academic backgrounds in computer science, ranging from advanced undergraduates to postgraduate students, to participate as human annotators. Each participant spent around five hours verifying scientific claims, evaluating the associated reasoning steps, and conducting error analysis based on tabular data.
Participation was entirely voluntary, with no financial incentives provided. The annotators volunteered intrinsically. To ensure annotation quality and fairness, we provided clear guidelines and a brief orientation before the task began. The consistency observed across participants' annotations supports the reliability and validity of the resulting dataset.

\subsection{Claim Annotation}

When annotating claims, if there are inconsistencies in the labels assigned across different reasoning paths (i.e., the multipath labels are not consistent), an additional round of manual review is conducted. In this step, a human annotator considers all the available information from previous annotation attempts and determines a final, consolidated label to ensure reliability and accuracy in the dataset.

\subsection{Skill-Chain Evaluation}
The skill-chain evaluation includes granularity, information redundancy, alignment, interpretability, and accuracy. Among these, accuracy, alignment, and redundancy are scored automatically using GPT-based evaluation. In contrast, granularity and interpretability require human judgment due to their subjective nature.

\section{Additional Details for Prompts}

\begin{figure*}[t]
\centering
\small
\begin{Case Study with Fixed-ratio Inappropriate Context}

SYS\_PROMPT = f"""\#\#\# **Task** 

You are a helpful assistant to give several claims that could be inferred by the table content. Please follow the steps below to give your answer:

1. Read the table content carefully and try to understand what information is given in the table.

2. Identify more than five key aspects that you can make claims about \(trend, maximum, average, inference, etc. \) that are meaningful in the domain. When writing the claim, ensure you incorporate specific knowledge from the field related to the table. Naturally incorporate the domain knowledge into the calculation. 

3. Make the claims complex in mathematical calculation but clear in expression. The data in the table must clearly support the claim based on physical principles of the domain or experimental facts, not just superficial correlations.

4. In order to verify the claim, commplex calculations like multi-step complex deduction, sum, trend, multiplication and etc. should be needed.

5. Adjust the claim to be more deterministic, precise, diverse, and complex. Delete vague words like "poorly", "smilarly", "substantially", "consistently" and "significantly". Change vague words to comparative metrics like "perform worse than", "same", and "increase" and include specific calculated numbers from the table.

6. Write the scientific claim to make it more natural by integrating the domain knowledge into the numerical trend rather than explicitly stating it. The revision should maintain a formal scientific tone, keep the focus on the numerical relationship, and avoid directly explaining the underlying mechanism. Convey the scientific conclusion implicitly through the data variation

7. The claim should involve complex and challenging calculations, requiring a deep understanding of the table as well as partial knowledge of the domain. Naturally incorporate the domain knowledge into the calculation. It goes beyond simple cell-to-cell operations or comparisons. Include multi-step implicit mathematical calculation in the claim and do not explictly write all the steps. The calculation process must include more than five steps, at most eight steps, and most of the steps need to be include in the implicit calculation. When generating this claim, the intermediate calculation steps should not be written out, and the specific numerical values from the table should not be mentioned. Only the final conclusion should be presented. 

8. Do not write claims that need to be verified by locating all the cells in the table. Generate claims that require calculation between several cells in the table. Avoid trivial numerical comparison. Involve complex multi-step implicit computation for the claim.

9. Check the calculation results to be correct, if it is not correct, calculate it again and ensure the final results shown in the claim is correct. The claim should include the final numerical computational result. Write concrete determinstric claim. Avoid speculative sentences like "possibly due to..." or "may..."

\#\#\# **Response Format**
Your Response:
\#\#\# Understand the Table
[thoughts about the table content]

\#\#\# Claim Aspects
[more than five aspects of the table content that you can make claims about]
[aspect 1], [aspect 2], [aspect 3], ...

\#\#\# Claims Details
[one claim about each aspect, each in a separate bullet point]
- [aspect 1]: [claim 1]
- [aspect 2]: [claim 2]
..."""

USER\_PROMPT = """\#\#\# Table
<caption>
<table>

Your Response:
"""

\end{Case Study with Fixed-ratio Inappropriate Context}
\caption{Data Augmentation Prompt before Claim Generation and Generate the Positive Claim.}
\end{figure*}

\begin{figure*}[t]
\centering
\small
\begin{Case Study with Fixed-ratio Inappropriate Context}
\#\#\# **Example**

\#\#\# Table
Caption
Amount of freezable water and non-frozen water in XLPE/silica nanocomposites conditioned at $50^{\circ}\mathrm{C}$ 100\% th from MDSC measurement (one sample for each material).

Table
| material | Freezable water (mg/g) | Non-frozen water (mg/g) | Total water (mg/g) |
| XLPE | CellTag | 0.4 | 0.4 |
| 5 wt\% VS | 1.1 | 2.6 | 3.7 |
| 12.5 wt\% VS | [BOLD] 5.3 | [BOLD] 7.7 | [BOLD] 13.0 |

\#\#\# Original Claim
The non-frozen water content also increases with higher silica content, and at a higher rate compared to freezable water, suggesting that silica's interaction with water molecules predominantly enhances the freezable fraction.

Your Response:
\#\#\# Thought
The original claim states that non-frozen water content increases with higher silica content, and at a higher rate compared to freezable water. To make the claim not supported by the table, I can alter the rate comparison, suggesting that non-frozen water increases at a slower rate than freezable water, which contradicts the data.

\#\#\# Claim
The non-frozen water content also increases with higher silica content, but at a slower rate compared to freezable water, suggesting that silica's interaction with water molecules predominantly enhances the freezable fraction.

\end{Case Study with Fixed-ratio Inappropriate Context}
\caption{Few Shot Examples Used When Performing Negative Claim Generation}
\end{figure*}

\begin{figure*}[t]
\centering
\small
\begin{Case Study with Fixed-ratio Inappropriate Context1}

EXAMPLE = {`interpret':'The claim is comparing the performance of two models trained using discriminative methods: FINE\-TUNED\-DISCRIMINATIVE and CS\-ONLY\-DISCRIMINATIVE, specifically focusing on their performance on the test set. The claim states that the FINE\-TUNED\-DISCRIMINATIVE model is superior to the CS\-ONLY\-DISCRIMINATIVE model in terms of test perplexity (test perp), test accuracy (test acc), and test word-error-rate (test wer). Here, "perp" refers to perplexity, "acc" to accuracy (measured in percent), and "wer" to word-error-rate. In general, a lower perplexity and word-error-rate indicate better model performance, while a higher accuracy indicates better performance.', 'plan':'[Plan 1 Start]Extract the \'test perp\' value for the \'CS-only-disc\' model, and the \'test perp\' value for the \'Fine-Tuned-disc\' model, and then compare these two values to verify if the \'test perp\' value of the \'Fine-Tuned-disc\' model is lower than the \'test perp\' value of the \'CS-only-disc\' model. [Plan 1 End]

[Plan 2 Start]Extract the \'test acc\' value for the \'CS-only-disc\' model, and the \'test acc\' value for the \'Fine-Tuned-disc\' model, and then compare these two values to verify if the \'test acc\' value of the \'Fine-Tuned-disc\' model is higher than the \'test acc\' value of the \'CS-only-disc\' model. [Plan 2 End]

[Plan 3 Start]Extract the \'test wer\' value for the \'CS-only-disc\' model, and the \'test wer\' value for the \'Fine-Tuned-disc\' model, and then compare these two values to verify if the \'test wer\' value of the \'Fine-Tuned-disc\' model is lower than the \'test wer\' value of the \'CS-only-disc\' model. [Plan 3 End]', 'cell':'To extract the \'test perp\' value for the \'CS-only-disc\' model, we first locate the row corresponding to \'CS-only-disc\'. Counting from the first row: 1. Spanish-only-LM row 2. English-only-LM row 3. All:CS-last-LM row 4. All:Shuffled-LM row 5. CS-only-LM row 6. CS-only+vocab-LM row 7. Fine-Tuned-LM row 8. CS-only-disc row. So the \'CS-only-disc\' model is in the 8th row. Next, we locate the column for \'test perp\'. Counting from the first column: 1. performance column 2. dev perp column 3. dev acc column 4. dev wer column 5. test perp column. So the \'test perp\' column is the 5th column. Then we locate the 5th column of the ‘CS-only-disc' row. The cell at the intersection of the 8th row and the 5th column is the \'test perp\' value for \'CS-only-disc\'. The value is \'1.3\'. To extract the \'test perp\' value for the \'Fine-Tuned-disc\' model, we first locate the row corresponding to \'Fine-Tuned-disc\'. Counting from the first row: 1.Spanish-only-LM row 2.English-only-LM row 3.All:CS-last-LM row 4.All:Shuffled-LM row 5.CS-only-LM row 6.CS-only+vocab-LM row 7.Fine-Tuned-LM row 8.CS-only-disc row 9.Fine-Tuned-disc row. So the \'Fine-Tuned-disc\' model is in the 9th row. The \'test perp\' column is the 5th column. Then we locate the 5th column of the \'Fine-Tuned-disc\' row. The cell at the intersection of the 9th row and the 5th column is the \'test perp\' value for \'Fine-Tuned-disc\'. The value is \'2.8\'.', 'extract': 'The \'test perp\' value for the \'CS-only-disc\' model is 1.3. The \'test perp\' value for the \'Fine-Tuned-disc\' model is 2.8.', 'reason': 'The \'test perp\' value for the \'CS-only-disc\' model is 1.3. The \'test perp\' value for the \'Fine-Tuned-disc\' model is 2.8. To verify if the \'test perp\' value of the \'Fine-Tuned-disc\' model is lower than the \'test perp\' value of the \'CS-only-disc\' model, we compare these two values. Comparing 2.8 and 1.3, we find that 2.8 is greater than 1.3. Therefore, the \'test perp\' value of the \'Fine-Tuned-disc\' model (2.8) is not lower than the \'test perp\' value of the \'CS-only-disc\' model (1.3). This indicates that based on the \'test perp\' metric, the Fine-Tuned-disc model does not outperform the CS-only-disc model. In fact, a lower perplexity indicates better performance, so the CS-only-disc model has a better \'test perp\' value than the Fine-Tuned-disc model.',
'recap':'Based on the comparison of \'test perp\' values, we found that the \'Fine-Tuned-disc\' model has a higher \'test perp\' value (2.8) than the \'CS-only-disc\' model (1.3), thus failing to support the claim that Fine-Tuned-disc model outperforms CS-only-disc model in terms of test perplexity as stated in Plan <plan\_idx>. Since the claim requires outperforming on all metrics and we have already found a contradiction in \'test perp\', the overall claim is false. <flag>False</flag>',
'conclusion': 'Based on the table, we examined the test perplexity, test accuracy, and test word\-error\-rate for both the CS\-ONLY\-DISCRIMINATIVE and FINE\-TUNED\-DISCRIMINATIVE models. We found that the FINE\-TUNED\-DISCRIMINATIVE model has a higher test perplexity (2.8) compared to the CS-ONLY-DISCRIMINATIVE model (1.3), indicating a worse performance in terms of perplexity. While the FINE-TUNED-DISCRIMINATIVE model shows better performance in test accuracy and test word-error-rate, outperforming the CS-ONLY-DISCRIMINATIVE model in these metrics, it fails to outperform in test perplexity. Therefore, the claim that FINE-TUNED-DISCRIMINATIVE modeling outperforms CS-ONLY-DISCRIMINATIVE model on test perplexity, test accuracy, and test word-error-rate is refuted by the table.'}

\end{Case Study with Fixed-ratio Inappropriate Context1}
\caption{Examples used during atomic reasoning}
\end{figure*}

\begin{figure*}[t]
\centering
\small
\begin{Case Study with Fixed-ratio Inappropriate Context1}
SYS\_PROMPT[`interpret']= f"""\#\#\# Task
(1. Understand the problem)
You are a helpful assistant to help me interpret a claim based on a table input. You are given a table and a claim which is based on the table. Now, please interpret the claim based on the content in the table, please follow the guidelines below.

\#\#\# Guidelines
Please only interpret the claim but do not give the answer or solution, just make sure you understand what you need to do. Make sure the interpretation is concise and clean.
Please solve any ambiguity or reference that may exist in the question, and give your interpretation only based on the caption, table, and claim.
Please reason comprehensively and be careful to consider all conditions, constraints and all possible meanings in your problem interpretation.

\#\#\# Example
Here is an example interpretation:
{EXAMPLE['interpret']}
"""

USER\_PROMPT['interpret']= """\#\#\# Table Content
<caption>

<table>

\#\#\# Claim
<claim>

\#\#\# Your Interpretation of Claim
<interpretation>
"""

\end{Case Study with Fixed-ratio Inappropriate Context1}
\caption{Atomic Reasoning Chain Generation: Step 1}
\end{figure*}

\begin{figure*}[t]
\centering
\small
\begin{Case Study with Fixed-ratio Inappropriate Context1}
SYS\_PROMPT[`plan'] = f"""\#\#\# Task
(2. Give a Plan)
You are a helpful assistant in giving a step-by-step plan based on the interpretation of a given claim. Your goal is to determine whether the claim is supported, refuted, or cannot be verified (not enough information) based solely on the information provided in the table. Please follow the guidelines below to give the plan.

\#\#\# Guidelines
Please list as concrete steps in the plan as possible. Each plan is to verify one subclaim of the whole claim based on the interpretation of the given claim. Make sure each step doesn't have an overlap. When conducting the planning, you should take into consideration of all the mentioned specialized condition in the claim and make plan to verify all the information of the claim.
Frame each plan only in one sentence in a clear, succinct way, avoiding any ambiguous or implicit reference, ensuring each plan including complete procedures for each subclaim verification step.
Please wrap your plan in [Plan 1 Start] … [Plan 1 End], [Plan 2 Start] … [Plan 2 End] format

\#\#\# Example
Here is an example plan:
{EXAMPLE['plan']}
"""

USER\_PROMPT['plan'] = """\#\#\# Table Content
<caption>

<table>

\#\#\# Claim
<claim>

\#\#\# Interpretation of Claim
<interpretation>

\#\#\# Your Plan
<plan>
"""

\end{Case Study with Fixed-ratio Inappropriate Context1}
\caption{Atomic Reasoning Chain Generation: Step 2}
\end{figure*}

\begin{figure*}[t]
\centering
\small
\begin{Case Study with Fixed-ratio Inappropriate Context1}

SYS\_PROMPT[`cell'] = f"""\#\#\# Task
(3. Ground the cell with information mentioned in subplan)
You are an expert table data extraction assistant. Your task is to precisely locate and extract specific cell values from the provided table based on the instructions given in the subplan. You must strictly follow the subplan and the guidelines provided below.

\#\#\# Guidelines
Your broader aim is to extract all the required information mentioned by the subplan. To achieve this, you need to ground each cell with required information. Currently, you are working on the <plan\_idx> step of the plan: [Plan <plan\_idx>].
Please only try to ground cells and extract data for the designated subplan step by step, DO NOT perform other steps!
When locating the cell, you always count from the first cell (head) of the columns or rows to locate the row corresponding to the entity mentioned in the subplan. You count row from the first row, count column from the first column. 
When locating cells and extracting data from the column or row, indicate the entities you need to locate and extract with sufficient steps.
You should output your grounding steps in <grounding>...</grounding> format, and output your sentences of the extracted data of the grounded cells in <extraction>...<\/extraction> format.

\#\#\# Example
Here is an example cell grounding and extraction:
<grounding>
{EXAMPLE[\'cell\']}
<\/grounding>

<extraction>
{EXAMPLE['extract']}
<\/extraction>
"""

USER\_PROMPT['cell'] = f"""\#\#\# Table Content
<caption>

<table>

\#\#\# Claim
<claim>

\#\#\# Subplan
<subplan>

\#\#\# Your Grounding and Extraction
<grounding>

<extraction>
"""

\end{Case Study with Fixed-ratio Inappropriate Context1}
\caption{Atomic Reasoning Chain Generation: Step 3}
\end{figure*}

\begin{figure*}[t]
\centering
\small
\begin{Case Study with Fixed-ratio Inappropriate Context1}
SYS\_PROMPT[`reason'] = f"""\#\#\# Task
(4. Give a Reasoning with Skills)
You are a helpful assistant to use your own knowledge and reasoning to implement a particular step in the plan, in order to verify the claim based on the table. Please follow the guidelines below when performing the reasoning on your subgoal.

\#\#\# Guidelines
Your broader aim is to verify whether the claim is supported, refuted based on the content of a table, or cannot be verified (not enough information) based solely on the information provided in the table. To achieve this, you have previously made a plan about how to achieve that goal, and you have the grounded cells containing the required data to verify the subplan. Currently, you are working on the <plan\_idx> step of the plan: [Plan <plan\_idx>].
Please only try to implement the reasoning of the designated plan step, DO NOT perform other steps!
To implement this step in the plan, you should first use the relevant information from the provided extracted cells in the table based on what you need. Then you should do the reasoning based on the extracted cell and data, including comparison, calculation, etc.
Please reason carefully, thoroughly, and coherently. Perform each step of reasoning with justification.

\#\#\# Example
Here is an example reasoning:
{EXAMPLE['reason']}
"""

USER\_PROMPT['reason'] = f"""\#\#\# Table Content
<caption>

<table>

\#\#\# Claim
<claim>

\#\#\# Subplan
<subplan>

\#\#\# Grounded Cell with Extracted Data
<grounding\&extraction>

\#\#\# Your Reasoning
<reasoning>
"""

\end{Case Study with Fixed-ratio Inappropriate Context1}
\caption{Atomic Reasoning Chain Generation: Step 4}
\end{figure*}

\begin{figure*}[t]
\centering
\small
\begin{Case Study with Fixed-ratio Inappropriate Context1}
SYS\_PROMPT[`recap'] = f"""\#\#\# Task
(5. Verify the Reasoning and Refer back to the Plan)
You are a helpful assistant in generating a coherent transition sentence to conclude what you have done in the previous reasoning step, refer to the whole plan, and look ahead about what to do next. Please follow the guidelines below to generate the transition sentence.

\#\#\# Guidelines
Your generated transition sentence should be coherent with previous reasoning content, and first logically conclude what result you get and whether you have achieved the goal of the subplan.
Then, you should refer back to the whole plan, see what you have done, and look ahead to what you should do next.
Please generate all these transitions within three sentences, keep your transition coherent, logical, and clear. 
If you find the subplan is verified to be false by your previous reasoning step, conclude why the subclaim is wrong and do not write the transition to next step, just conclude the whole claim is false since the subclaim is verified to be false, then give an ending flag formatted as <flag>Flase</flag>. If you find the subplan is verified to be true by your previous reasoning step, cleanly conclude how you verify it to be true, and give an ending flag formatted as <flag>True</flag>. If you find that the subplan can not be verified as either true or false with all the existing information, cleanly conclude what you have done for this subplan.

\#\#\# Here is an example transition:
{EXAMPLE['recap']}
"""

USER\_PROMPT['recap'] = f"""\#\#\# Table Content
<caption>

<table>

\#\#\# Claim
<claim>

\#\#\# All Plans
You are trying to verify this claim based on the content from the given table:
This is the whole plan of what you should do in order to verify this claim:
<plan>

\#\#\# Subplan
<subplan>

\#\#\# Reasoning
<reasoning>

\#\#\# Your transition
<transition>
"""

\end{Case Study with Fixed-ratio Inappropriate Context1}
\caption{Atomic Reasoning Chain Generation: Step 5}
\end{figure*}

\begin{figure*}[t]
\centering
\small
\begin{Case Study with Fixed-ratio Inappropriate Context1}
SYS\_PROMPT[`conclusion'] = f"""\#\#\# Task
(6. Conclude and Get Final Result)
You are a helpful agent to conclude the whole reasoning process and give your final response on whether the given claim is supported or refuted by the table or not enough information to verify it based on the information provided by the table. Please give your conclusion following the guidelines below.

\#\#\# Guidelines
You should first recap what you have achieved based on all previous reasoning steps.
You should consider carefully whether there are outliers or cases that you haven't taken into consideration, and whether they are important to your final conclusion.
After thinking thoroughly over all the information you get from previous reasoning steps and the table information, you should give your final response about whether the claim is supported or refuted by the table, or whether there is not enough information. Ensure all your sentences are based on the true information.
Please wrap your final answer in <conclusion></conclusion>.
If you find the claim is verified to be false by your previous steps, give an ending flag formatted as <flag>Flase</flag>. If you find the claim is verified to be true by your previous steps, give an ending flag formatted as <flag>True</flag>. If you find the claim can not be verified to false or true merely based on the information provided by the table, give an ending flag formatted as <flag>Not enough information</flag>.
You can give only one ending flag in the conclusion as the label for the claim.
Please give your conclusion within four sentences.

\#\#\# Here is an example conclusion:
{EXAMPLE['conclusion']}
"""

USER\_PROMPT['conclusion'] = f"""\#\#\# Table Content
<caption>

<table>

\#\#\# Claim
<claim>

\#\#\# Plans
<plan>

\#\#\# Reasoning and Transition
<allReasonTransition>

\#\#\# Your Conclusion
<conclusion>
"""

\end{Case Study with Fixed-ratio Inappropriate Context1}
\caption{Atomic Reasoning Chain Generation: Step 6}
\end{figure*}


\end{document}